# An Investigation in Optimal Encoding of Protein Primary Sequence for Structure Prediction by Artificial Neural Networks

Aaron Hein
Computer Science & Engineering
University of South Carolina
Columbia, SC, USA
ahein@email.sc.edu

Casey A. Cole
Computer Science & Engineering
University of South Carolina
Columbia, SC, USA
colca@email.sc.edu

Homayoun Valafar
Computer Science & Engineering
University of South Carolina
Columbia, SC, USA
homayoun@cse.sc.edu

**Abstract –** *Machine learning and the use of neural networks has increased precipitously over the past few years primarily due to the ever-increasing accessibility to data and the growth of computation power. It has become increasingly easy to harness the power of machine learning for predictive tasks. Protein structure prediction is one area where neural networks are becoming increasingly popular and successful. Although very powerful, the use of ANN require selection of most appropriate input/output encoding, architecture, and class to produce the optimal results. In this investigation we have explored and evaluated the effect of several conventional and newly proposed input encodings and selected an optimal architecture. We considered 11 variations of input encoding, 11 alternative window sizes, and 7 different architectures. In total, we evaluated 2,541 permutations in application to the training and testing of more than 10,000 protein structures over the course of 3 months. Our investigations concluded that one-hot encoding, the use of LSTMs, and window sizes of 9, 11, and 15 produce the optimal outcome. Through this optimization, we were able to improve the quality of protein structure prediction by predicting the φ dihedrals to within 14° - 16° and ψ dihedrals to within 23°- 25°. This is a notable improvement compared to previously similar investigations.*

**Keywords*: Protein Folding, Neural Networks, LSTM, Torsion Angle Prediction, Tertiary Structure.***

## I. INTRODUCTION

Proteins are variable length chains of amino acid residues. Anfinsen's dogma states that simply the sequence of amino acid residues is enough to determine the unique, three-dimensional shape of a given protein [1]. However, in practice determining that structure purely from the first principles of physics and thermodynamics is computationally intractable especially for challenging proteins such as those that undergo dynamics or membrane proteins. Nevertheless, determining the structure is vital in determining the function of the protein. Perturbations in the structure of a protein can lead to a misfolding of the protein, which can lead to the manifestation of diseases. Alzheimer's disease, Type 2 Diabetes Mellitus and Parkinson's disease [2] can all be cited as examples of this phenomenon. In all of those cases, proteins fail to fold ”properly” causing a disruption in the natural cellular functions. It stands to reason that the cure to these diseases would involve drugs or therapies to correct the misfolded proteins.

Although there are experimental methods to determine protein structure [3] [4], these methods are highly time and cost intensive. In contrast, computational approaches to protein structure determination have many advantages such as reduced cost, increased effectiveness, and speed. Computational approaches can also be conducted in the absence of the biological sample. Although purely physics-based simulation of protein folding is intractable at this time, the recent advances in big data and machine learning have given rise to inception of data driven strategies to protein structure determination.

Previous works in this field have detailed the use of various machine learning techniques in order to make these predictions. These techniques have included support vector machines [5] [6] as well as a number of different neural network architectures [7] [8] [9] [10] [11] [12] [13]. Due to their obvious advantages, the most successful approaches to structure prediction of proteins have consisted of Artificial Neural Networks (ANN). Although ANNs provide several advantages over other existing Machine Learning techniques, they do require optimization in numerous categories including selection of a proper model of ANN, ANN architecture, and input/output encoding. Nearly all of the previous reports [7] [8] [9] [10] [11] [12] have consisted of an investigation of different architectures and models of ANN to improve performance. While the selection of the most appropriate model and architecture is an important aspect of an ANN-based approach, the proper selection of input and output encoding scheme has been poorly investigated despite its impact on the problem outcome.

Here we present an investigation and evaluation of multiple input encoding including schemes based off of 10 different substitution matrices and 11 different window sizes, while presenting a new approach that improves ANN's predictions compared to the previous approaches when tested on a single class of proteins. More specifically, we have utilized protein CATH class 1.10.510.10 that consists of 5,814 protein structures in order to develop and evaluate different input encoding schemes. We have tested general improvements of our new proposed encoding scheme on 7 models of different architectures as well as two other protein CATH classes, 2.60.120.200 and 3.90.1150.10, that collectively encompassed 4505 protein structures. In general, we observed a 4°-5°improvement in the prediction of the both torsion angles across different ANN models and architectures.

# II. BACKGROUND AND METHOD

## A. *Protein Structure Formulation in Rotamer Space*

An amino acid is an organic compound that is made up of an amine group (consisting of a nitrogen and two hydrogens) and a carboxyl group (consisting of a carbon with two oxygen's and a hydrogen) connected by a carbon atom. This central carbon is known as the alpha carbon ($C_\alpha$). The $C_\alpha$ atom also has a hydrogen bonded to it in addition to a side chain. This side chain is different for each amino acid. For example, Glycine has a very simple side chain consisting of a single hydrogen atom whereas Aspartate's side chain consists of a Carbon with two Oxygen atoms. A protein is a series of amino acids that have bonded together in what is known as a peptide bond ~~(shown in Figure 1).~~ During a peptide bond, the Carbon from the carboxyl group of the first amino acid bonds with the Nitrogen from the amine group of the second amino acid. Additionally, the Oxygen-Hydrogen from the carboxyl group of the first amino acid bond with a Hydrogen from the amine group of the second amino acid to form water ($H_2O$) which is released. The resulting structure's backbone consists of N – $C_\alpha$ – C – N – $C_\alpha$ – C. When this bond occurs, the angle that describes the rotation between the two amino acid residues is known as $\omega$ (omega). It is almost always fixed at 180 degrees. However, the other angles in the structure are much less rigid. It is these angles that define the protein's secondary and tertiary structure. The angle that describes the rotation around the bond between the Nitrogen and alpha Carbon is referred to as the $\varphi$ (phi) angle, while the angle that describes the rotation around the bond between the alpha Carbon and the carboxyl Carbon is known as the $\psi$ (psi) angle.

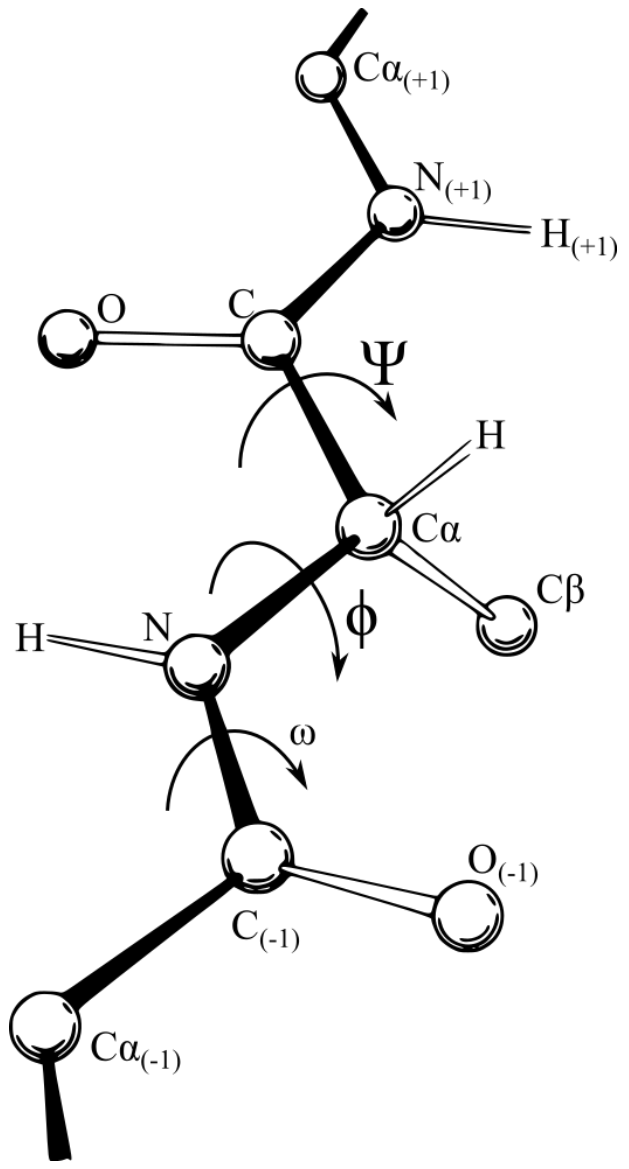

*Figure 1: An illustration of protein backbone dihedrals [14].*

## B. *Methodology*

All previous approaches to protein structure prediction using machine learning have striven to achieve more accurate predictions of the backbone dihedral angles φ (phi) and ψ (psi) for each amino acid. As early as 1988, Quin and Sejnowski [7] used machine learning and neural networks to predict these angles. In this earliest attempt, a window size of 13 amino acids and one-hot encoding was used to represent the data for input to the ANN. Since then, many others have used machine learning and neural networks to attempt more accurately prediction of these torsion angles. Reporting in chronological order: in 2005, Wood and Hirst proposed a method for predicting secondary structures and the ψ angles called DESTRUCT [8]. This method relied on the Position Specific Scoring Matrix (PSSM) from PSI-BLAST[15] as well as the φ angle as input data for their models. PSI-BLAST uses the BLOSUM62 scoring matrix [16] as the starting point by default, although other matrices can be used which will impact the encoding of the protein. Additional use of neural networks to predict the torsion angles were made in the ANGLOR[9], REALSPINE 2.0 [17], REALSPINE 3.0 [18], SPINE XI [19], SPINE X [20], SPIDER 2 [11], Raptor-X [10] and deep learning methods works[12] with all of them including the PSSM from PSI-BLAST in their input features.

### C. *The Explored Encoding Schemes*

Our exploration of the new encoding schemes was motivated by three constraints. First was to avoid dependence on an encoding mechanism that is time-variant such as the use of PSSM that is continuously changing due to deposition of new proteins. Second was to develop a method that did not require a heavy pre-processing step. The final constraint was to develop an encoding mechanism that will broadly improve structure prediction across different protein classes, models, and architectures of ANN. To that end, we used a one-hot encoding as well as ten other encoding schemes based on different BLOSUM [21] and PAM [22] substitution matrices including BLOSUM62 [16] (used in PSSM creation by PSI-BLAST), BLOSUM30, BLOSUM45, BLOSUM65, BLOSUM80, BLOSUM 100, PAM30, PAM60, PAM120 and PAM250. A twenty-one position array was used for the encoding with the last position being reserved for other or unknown amino acid residues. As the name implies, one hot encoding involves creating a zero-filled array where each column represents one particular value. To represent that value, the column is set to one while all other entries remain zero. For example, in one-hot encoding (which has been commonly used in previous works), Alanine is represented by a column in a twenty-one unit vector. We used the same ordering across all of the encodings which was A R N D C Q E G H I L K M F P S T W Y V X. Therefore, ALA would be converted to [1, 0, 0, 0, 0, 0, 0, 0, 0, 0, 0, 0, 0, 0, 0, 0, 0, 0, 0, 0, 0]. With the BLOSUM62 encoding, the row corresponding to Alanine would be used resulting in the encoding [4, -1, -2, -2, 0, -1, -1, 0, -2, -1, -1, -1, -1, -2, -1, 1, 0, -3, -2, 0, 0].

Using our amino acid encoding, the complete proteins were then presented to the ANN using a sliding window of an odd number of amino acids. We explored window sizes ranging from 3 to 23 residues with the aim of predicting the torsion angles of the middle residue. After the residue was encoded, the $\varphi$ and $\psi$ angles of all residues were encoded using the sine and cosine functions as described in the Raptor-X [10] and deep learning works [12]. For each angle the sine and cosine were evaluated and used as the predicted outputs during the training sessions. This had the benefit of compressing the values between -1 and 1 which maps directly to the output of the $tanh()$ activation function while also allowing us to unambiguously retrieve the original angle using the $arctan2()$ function.

### D. *Target Proteins*

In our investigations, we used three protein classes (shown in Table 1) reported by the CATH classification mechanism as training data. CATH [23] is a method of classifying proteins by [C]lass, [A]rchitecture, [T]opology and [H]omologous superfamily. We focused primary on the 4 Classes at the top of the CATH hierarchy. They are mainly alpha (1), mainly beta (2), alpha beta (3), and few secondary structures (4). We chose to exclude class 4 for two reasons. First it is the smallest Class with only 4519 domains. Limited data makes it more difficult to train a neural network and to provide a meaningful review of the encoding and window sizes. Second, the catch-all nature of the class indicates much more diverse proteins that may be difficult or impossible for a neural network to learn during training.

*Table 1: A summary of the structures and CATH classes that were used in this experiment.*

| CATH Class | Domains | Unique PDBs |
|---|---|---|
| 1.10.510.10 | 5814 | 4042 |
| 2.60.120.200 | 2284 | 965 |
| 3.90.1150.10 | 2221 | 919 |

In total, these three classes represented more than 10,000 proteins with structural sampling of $\alpha$-helical, $\beta$-sheet, and mixed $\alpha/\beta$ proteins. In order to establish the generalization principle of our developed encoding, we used the proteins in Class 1 (1.10.510.10) to select the optimal encoding mechanism and used Classes 2 and 3 to validate the performance of our selected encoding.

### E. *Data Processing*

Prediction of the secondary or tertiary protein structure practically involves predicting the $\varphi$ (phi) and $\psi$ (psi) angles which are located around the $C_\alpha$ atom of each residue. These angles are commonly referred to as the torsion angles of the amino acid. In order to produce the $\varphi$ and $\psi$ angles, we made use of the online PDBMine tool [24]. The PDBMine database includes the backbone torsion angles for over 140,000 proteins and is accessible via an online web portal as well as via direct API calls. The content of PDBMine can be used for retrieval of protein torsion angles, or for data-mining purposes that can assist in the folding of proteins [24]. In this work, the use of PDBMine eliminated the need to process and store large amounts of data while also improving overall repeatability. One caveat of note is that PDBMine can have multiple models or versions of the same protein because multiple models can exist in the PDB. We were concerned that using all of them would skew the data set and give more weight to those proteins. Therefore, then there were multiple models of a protein in the PDB, we used the first model that was returned and discarded the others.

### F. *Artificial Neural Network Architectures*

Although the goal of this paper is not to develop or identify the best model but rather to identify the most suitable encoding and window size, it was required to try multiple models with the above encodings and window sizes in order to compare their initial performance. We explored different particularization of our input encoding on Feed Forward and LSTM models of neural networks. For each CATH data set, we explored different architectures as shown in Table 2. There is some intuition that an LSTM [25] might perform well when predicting the torsion angles due to the sequential nature of the data. However, for completeness, other model architectures were also used. We trained each data set on two 'regular' feed forward neural networks and on five LSTM networks. Each neural network architecture contained a different number of layers but the other training parameters remained constant. In between all of the dense layers, dropout regularization [26] was used with a 30% dropout rate.

*Table 2: A summary of ANN architectures investigated in this study.*

| Model | Architecture |
|---|---|
| DNN 1 | 3 layer feed forward neural network |
| DNN 2 | 6 layer feed forward neural network |
| LSTM 1 | 1 LSTM layer with 2 fully connected layers before output |
| LSTM 2 | 2 LSTM layer with 2 fully connected layers before output |
| LSTM 3 | 4 LSTM layer with 2 fully connected layers before output |
| LSTM 4 | 8 LSTM layer with 2 fully connected layers before output |
| LSTM 5 | 64 LSTM layer with 2 fully connected layers before output |

All of the models were trained with an initial learning rate of .01 which was set to reduce if learning plateaued during training using mean squared error as the loss function. Although many of the papers cited in this work used Mean Absolute Error as defined in equation 1, it can be argued that Mean Squared Error as defined in equation 2 is a better metric for training in this particular instance. RMSE gives more weight to larger errors since the errors are squared prior to summation. This seemed preferable as a large error would more drastically affect the structure of the protein.

$$MAE = \frac{1}{n}\sum_{i=1}^{n}|y_i - \hat{y}_i| \tag{1}$$

$$RMSE = \sqrt{\frac{1}{n}\sum_{i=1}^{n}(y_i - \hat{y}_i)^2} \tag{2}$$

A total of 2,541 models were investigated as part of this survey. Training on a CPU led to an estimated completion time of about 9 months. In order to decrease training time, a GPU was utilized. Each model was trained for 50 epochs and with a batch size of 4096 in an attempt to maximize GPU utilization. Training with the GPU lowered the initial estimated training time from 265 days with a single CPU to only 45 days. The training process was also configured to stop once the validation loss plateaued and the models were no longer benefiting from additional training. All of the layers used the ReLU activation function with the exception of the output layers which all used the tanh activation function.

### *G. Training, Testing, and Evaluation Protocols*

Three models were trained for each model architecture, window size and encoding combination. One was trained to predict only the $\varphi$ angle. A second was trained to predict only the $\psi$ angle, and a third was trained to predict both angles simultaneously. The thought was that perhaps information about both angles might lead to more accurate predictions.

Rather than encoding the data and then splitting into training, testing and validation sets, we randomly divided the list of proteins into a 70/20/10 split for training, validation and testing. The data was then encoded after the split. This ensured that the entire protein was included in the same data set although there was a small impact to the final percentages. In other words, because protein can be of a different length, when encoded it will produce a different number of samples. A protein that is 20 residues long when encoded with a window size of 5 will produce 16 samples whereas a protein that is 30 residues long will produce 26 samples. If the 30 residue protein was included in the training set and the 20 residue protein was included in the validation set, the training set would have a few more samples. Although the training set included 70% of the proteins, it was not exactly 70% of the samples. The models then were trained using a training set that was made up of 70% of the data and a validation set that was 20% of the data. 10% of the data was held out as a testing set and not used for training at all.

In order to gauge the effectiveness of just using the encodings, the predictions were then converted back into angles in degrees and the mean absolute error was calculated based on the angles. The lower the error, the better the model. This allowed us to compare these results with those achieved in the previous works. After the error results were compared and the best model architecture, encoding and window size identified, additional models were trained on the additional two CATH classes. Because CATH Class 1 is made up of mostly $\alpha$-helical structures, we wanted to ensure that our findings were applicable across proteins with different secondary structures, so we selected two additional CATH classes to use as additional data sets. This provided verification that the data processing has the potential to be expanded across all proteins in the future. For this additional validation, separate models were trained using CATH 2.60.120.200 and then trained again using CATH 3.90.1150.10. The only difference from the initial training was that they were trained for 150 epochs as opposed to only 50 epochs but with early stopping still applied. This was done in an attempt to improve the results and further validate that the results were good enough to pursue additional research using this data preparation. In order to make an accurate comparison, CATH 1.10.510.10 was used again to train with 150 epochs on the narrowed selection.

## III. RESULTS AND DISCUSSION

### A. *Root Mean Squared Error versus Mean Absolute Error*

In order to evaluate model performance, we examined the error rate when making predictions on the previously discussed testing set. Because the Root Mean Squared Error was used for training, we looked at that first. Tables 3, 4, and 5 show the twenty lowest RMSE achieved across all 2,541 models trained as well as the encoding, window size and architecture that produced the lowest error. Specifically, Table 3 shows the twenty models with the lowest RMSE for predicting the φ angles. Table 4 shows the models with the lowest RMSE for predicting the ψ angles. The models with the lowest RMSE for predicting both the φ and ψ angles at the same time are in Table 5.

*Table 3: Lowest MSE for Phi prediction.*

| Score | Encoding | Window Size | Model Arch |
|---|---|---|---|
| 0.062 | one hot | 9 | lstm 5 |
| 0.062 | one hot | 11 | lstm 5 |
| 0.063 | one hot | 13 | lstm 5 |
| 0.064 | one hot | 15 | lstm 5 |
| 0.065 | one hot | 7 | lstm 5 |
| 0.068 | blosum30 | 7 | lstm 5 |
| 0.068 | blosum45 | 7 | lstm 5 |
| 0.069 | one hot | 5 | lstm 5 |
| 0.069 | one hot | 17 | lstm 5 |
| 0.069 | blosum62 | 7 | lstm 5 |
| 0.069 | blosum65 | 7 | lstm 5 |
| 0.071 | one hot | 19 | lstm 5 |
| 0.071 | blosum45 | 5 | lstm 5 |
| 0.073 | blosum62 | 5 | lstm 5 |
| 0.073 | blosum65 | 5 | lstm 5 |
| 0.075 | pam250 | 5 | lstm 5 |
| 0.075 | blosum80 | 7 | lstm 5 |
| 0.076 | pam250 | 7 | lstm 5 |
| 0.076 | blosum65 | 9 | lstm 5 |
| 0.077 | pam120 | 7 | lstm 5 |

*Table 4: Lowest MSE for Psi prediction.*

| Score | Encoding | Window Size | Model Arch |
|---|---|---|---|
| 0.097 | one hot | 19 | lstm 5 |
| 0.098 | one hot | 13 | lstm 5 |
| 0.103 | one hot | 11 | lstm 5 |
| 0.103 | one hot | 15 | lstm 5 |
| 0.104 | one hot | 9 | lstm 5 |
| 0.107 | blosum62 | 7 | lstm 5 |
| 0.112 | one hot | 7 | lstm 5 |
| 0.113 | blosum30 | 7 | lstm 5 |
| 0.113 | blosum30 | 9 | lstm 5 |
| 0.113 | blosum62 | 9 | lstm 5 |
| 0.115 | blosum30 | 11 | lstm 5 |
| 0.116 | blosum45 | 9 | lstm 5 |
| 0.117 | blosum45 | 13 | lstm 5 |
| 0.119 | blosum45 | 7 | lstm 5 |
| 0.119 | blosum65 | 11 | lstm 5 |
| 0.12 | pam250 | 7 | lstm 5 |
| 0.12 | blosum45 | 11 | lstm 5 |
| 0.121 | blosum45 | 5 | lstm 5 |
| 0.123 | pam250 | 11 | lstm 5 |
| 0.124 | pam250 | 9 | lstm 5 |

*Table 5. Lowest MSE for Phi and Psi prediction.*

| Score | Encoding | Window Size | Model Arch |
|---|---|---|---|
| 0.097 | one hot | 11 | lstm 5 |
| 0.103 | one hot | 15 | lstm 5 |
| 0.104 | one hot | 17 | lstm 5 |
| 0.107 | one hot | 7 | lstm 5 |
| 0.107 | one hot | 9 | lstm 5 |
| 0.111 | blosum30 | 9 | lstm 5 |
| 0.111 | blosum45 | 7 | lstm 5 |
| 0.111 | blosum62 | 9 | lstm 5 |
| 0.113 | blosum62 | 7 | lstm 5 |
| 0.113 | blosum65 | 7 | lstm 5 |
| 0.114 | one hot | 13 | lstm 5 |
| 0.114 | blosum30 | 7 | lstm 5 |
| 0.114 | blosum65 | 11 | lstm 5 |
| 0.115 | blosum45 | 9 | lstm 5 |
| 0.116 | blosum80 | 9 | lstm 5 |
| 0.119 | pam250 | 7 | lstm 5 |
| 0.119 | blosum45 | 11 | lstm 5 |
| 0.12 | pam250 | 9 | lstm 5 |
| 0.121 | one hot | 5 | lstm 5 |
| 0.122 | pam250 | 5 | lstm 5 |

Because much of the past work in this field has relied on the Mean Absolute Error, we believed it important to look at that as well in order to make comparisons. Tables 6, 7 and 8 show the lowest MAE error rates for $\varphi$, $\psi$ and both $\varphi$ and $\psi$ angles respectively. It is worth pointing out that the RMSE will always have a value greater than or equal to the MAE. Instances where a model has a low MAE but a higher RMSE respective to other models indicates that when mistakes were made in the predictions they're generally larger mistakes. However, in our case, when training we're looking at the RMSE of the prediction which is of the sine and cosine of the angles and that was what we included in the tables. When we calculate the MAE, we're calculating the MAE of the actual angles. That means that the MAE will look larger because it's given in angle degrees and not the sine and the cosine.

*Table 6: Lowest angle MAE for Phi prediction.*

| Score | Encoding | Window Size | Model Arch |
|---|---|---|---|
| 14.294 | one hot | 9 | lstm 5 |
| 14.313 | one hot | 11 | lstm 5 |
| 14.421 | one hot | 13 | lstm 5 |
| 14.433 | one hot | 7 | lstm 5 |
| 14.571 | one hot | 15 | lstm 5 |
| 15.087 | blosum65 | 7 | lstm 5 |
| 15.133 | blosum45 | 7 | lstm 5 |
| 15.278 | one hot | 5 | lstm 5 |
| 15.29 | blosum62 | 7 | lstm 5 |
| 15.591 | blosum30 | 7 | lstm 5 |
| 15.615 | one hot | 17 | lstm 5 |
| 15.758 | one hot | 19 | lstm 5 |
| 15.823 | blosum62 | 5 | lstm 5 |
| 15.883 | blosum65 | 5 | lstm 5 |
| 15.915 | blosum45 | 5 | lstm 5 |
| 16.057 | pam250 | 5 | lstm 5 |
| 16.506 | blosum30 | 9 | lstm 5 |
| 16.522 | blosum30 | 5 | lstm 5 |
| 16.553 | blosum65 | 9 | lstm 5 |
| 16.574 | pam250 | 7 | lstm 5 |

*Table 7: Lowest angle MAE for Psi prediction.*

| Score | Encoding | Window Size | Model Arch |
|---|---|---|---|
| 23.289 | one hot | 19 | lstm 5 |
| 24.864 | one hot | 13 | lstm 5 |
| 25.227 | one hot | 11 | lstm 5 |
| 25.257 | one hot | 15 | lstm 5 |
| 25.335 | one hot | 9 | lstm 5 |
| 25.88 | blosum62 | 7 | lstm 5 |
| 26.313 | one hot | 7 | lstm 5 |
| 26.357 | blosum30 | 9 | lstm 5 |
| 26.408 | blosum45 | 5 | lstm 5 |
| 26.455 | blosum62 | 9 | lstm 5 |
| 26.465 | blosum30 | 7 | lstm 5 |
| 26.617 | blosum45 | 9 | lstm 5 |
| 26.655 | blosum30 | 11 | lstm 5 |
| 26.869 | blosum45 | 13 | lstm 5 |
| 27.056 | blosum65 | 11 | lstm 5 |
| 27.161 | blosum45 | 7 | lstm 5 |
| 27.378 | pam250 | 7 | lstm 5 |
| 27.47 | blosum45 | 11 | lstm 5 |
| 27.513 | pam250 | 9 | lstm 5 |
| 27.561 | pam250 | 11 | lstm 5 |

*Table 8: Lowest angle MAE for Phi and Psi prediction.*

| Score | Encoding | Window Size | Model Arch |
|---|---|---|---|
| 22.616 | one hot | 11 | lstm 5 |
| 23.229 | one hot | 9 | lstm 5 |
| 23.249 | one hot | 17 | lstm 5 |
| 23.586 | one hot | 15 | lstm 5 |
| 23.994 | one hot | 7 | lstm 5 |
| 24.067 | blosum62 | 9 | lstm 5 |
| 24.224 | blosum30 | 9 | lstm 5 |
| 24.371 | one hot | 13 | lstm 5 |
| 24.483 | blosum45 | 7 | lstm 5 |
| 24.515 | blosum62 | 7 | lstm 5 |
| 24.532 | blosum30 | 7 | lstm 5 |
| 24.782 | blosum65 | 11 | lstm 5 |
| 24.924 | blosum45 | 9 | lstm 5 |
| 25.132 | blosum45 | 5 | lstm 5 |
| 25.148 | blosum65 | 7 | lstm 5 |
| 25.166 | blosum45 | 11 | lstm 5 |
| 25.173 | blosum80 | 9 | lstm 5 |
| 25.215 | one hot | 5 | lstm 5 |
| 25.328 | pam250 | 9 | lstm 5 |
| 25.614 | pam250 | 5 | lstm 5 |

## B. *Optimal Architecture*

Although our goal was not to determine an optimal architecture, we did examine all of our explored neural network architectures to determine if any of the architectures performed better than the others. Interestingly, all of the best performing models were the deep (64 layer) LSTM model. This indicates that our intuition about an LSTM being a good choice for these types or predictions was correct. This is most likely due to the LSTM's strengths when looking at data in a series as well as the fact that this was the deepest network trained.

## C. *Encoding and Window Size*

When analyzing the best performing models, we were surprised to see that regardless of whether the model was trained to predict $\varphi$, $\psi$ or both, the five models with the lowest error were all trained using the one-hot encoded data. In fact, eight out of the twenty models with the lowest error when predicting only the $\varphi$ angle used one-hot encoding as well as six out of the twenty models with the lowest error when predicting only the $\psi$ angle. The next best encoding was BLOSUM45 which accounted for two of the twenty $\varphi$ models and five of the twenty $\psi$ models. This seems to indicate that one-hot encoding is a perfectly acceptable method of encoding the amino acids for use in training these types of models. This went against our intuition that one-hot encoding would serve as a baseline for comparison but would not perform as well as the encodings based off of substitution matrices because of the number of 0's in the encoding. We believed this would be the case because neural networks are basically a series of multiplication of a weight and the input value and anything multiplied by 0 equals 0. This can lead to difficulty training where there are a large number of 0's as is the case in one-hot encoding. Since one-hot encoding performed so well here, we have to assume that the additional relationship information we believed the substitution matrices to contain was not useful to the neural network.

The window size for the best performers was not nearly as clear as the best type of encoding was. Although the model with the lowest error when predicting $\varphi$ used a window size of 9, only two of the twenty models with the lowest error were training on data with a window size of 9 as opposed to eight of the twenty models that were trained on data with a window size of 7. Compare this with the models that were predicting the $\psi$ angles where five of the twenty models with the lowest error used a window size of 7, five used a window size of 9 and five used a window size of 11. The correlation seems to be that $\varphi$ predictions did better with a window size of 5 or 7 and $\psi$ predictions did better with a larger window size of 7, 9 or 11. When attempting to predict both the $\varphi$ and $\psi$ angles simultaneously, a window size of 7 and 9 was most consistently accurate with six models each out of the twenty models with the lowest error. Our conclusion is that 7 is the best window size overall for training models with these predictions. It's worth noting that no models with a window size of 3, 21 or 23 were among the best performers.

### *D. Expanded CATH Selection*

As previously discussed, after identifying the best model, encoding and window sizes, new models were trained using CATH 1.10.510.10, CATH 2.60.120.200 and CATH 3.90.1150.10 to ensure that similar results were achieved with proteins from different CATH classifications. This more focused retraining used only the deep LSTM (64 layer) architecture with one-hot encoding and more limited window sizes of 7, 9, 11, and 13. Each model was trained for 150 epochs, but with the same early stopping criteria. After training, the results were very similar to the first models. The results for the model trained on CATH 1.10.510.10 changed very little with the addition of more epochs. Those differences can be attributed to the nondeterministic nature of model training. For example, dropout regularization chooses nodes to ignore at random. This could result in different nodes being ignored in a different training. The results for the model trained on CATH 2.60.120.200 performed about 2 degrees worse for the $\varphi$ angle prediction and almost identically for the $\psi$ angles. Finally for the model trained on CATH 3.90.1150.10 performed about 1 degree worse than the original model. This could be because CATH 2 contains mostly $\beta$-sheets and the $\varphi$ and $\psi$ angles for this class of proteins are a little more difficult to predict than the $\varphi$ and $\psi$ angles for the mostly $\alpha$-helical proteins. Regardless, the results show great promise compared to the results found in previous works.

### *E. Overall Performance*

Spider 2 [11] achieved a MAE of 19° for $\varphi$ angle prediction and 30° for $\psi$ angle prediction. Raptor X [10] improved on that by 0.5°for $\varphi$ and 1.4°for $\psi$. We achieved a MAE of 14° - 16°for $\varphi$ angle prediction and 23°- 25°for $\psi$ angle prediction depending on the CATH class predicted. However, our results were limited to predictions over proteins in specific CATHs whereas the past work used proteins across a much wider spectrum. As the goal of this work was survey encoding and window size options and to identify an encoding, window size and neural network architecture for future work, these results seem very promising.

## IV. CONCLUSIONS AND FUTURE WORK

One of the biggest opportunities for future work lies in attempting to tune the neural network architecture to achieve better results. We did not focus on the network architecture or attempt to fine tune the hyper parameters. As such there are multiple potential improvements to be made to the model architecture to improve the predictions. Additionally, this work was performed on only three CATHs, but ideally the predictions work for any CATH and without the protein being classified beforehand. There are multiple ways this might be accomplished which presents its own avenue of research.

## V. ACKNOWLEDGMENT

We would like to thank Chris Ott for modifying the PDBMine interface. Funding for this work was provided by NIH Grant Number P20 RR-016461 to Dr. Homayoun Valafar.